\documentclass{article} %
\ifdefined\pdfsuppressptexinfo
    \pdftrailerid{}
\fi
\usepackage{times}

\usepackage{amsmath}

\usepackage[hyperfootnotes=false]{hyperref}
\usepackage{url}
\usepackage{graphicx}
\usepackage{float}
\usepackage{wrapfig}
\usepackage{needspace}
\usepackage{booktabs}
\usepackage{colortbl}
\usepackage{amssymb}
\usepackage{multirow}
\usepackage{tabularx}
\usepackage{siunitx}

\usepackage{latex/iclr2027_conference}

\title{Improving OCR Faithfulness via Gated and \mbox{Attenuated} On-Policy Distillation}

\author{Baode Wang\thanks{Equal contribution.} \quad Zuming Huang\footnotemark[1] \quad Kexuan Ren \quad Jun Huang \quad Wei Chu}

\begin{document}
\raggedbottom
\maketitle

\begin{abstract}
Vision-language models may rewrite anomalous text in images into linguistically plausible expressions, compromising OCR transcription faithfulness. Sequence-level task rewards and local teacher guidance are complementary, but guidance from the same teacher may not remain equally effective as the student improves. Offline analysis shows that supervision from a fixed teacher becomes progressively less favorable as the student improves, both across training checkpoints and across response groups with different task rewards. Motivated by this observation, we introduce \textbf{GAD-RL}, which adaptively regulates teacher supervision during joint post-training according to the student's current task performance and local distributions. A frozen teacher conditions on reference transcriptions and student-generated prefixes. GAD-RL disables distillation for response groups containing an output with task reward at least 0.95 and continuously attenuates distillation strength as group-mean reward increases. It also weights forward KL by the student's probability of the teacher's Top-1 token, moderating local auxiliary updates when student support for that candidate is low. On Qwen3.5-2B, GAD-RL achieves 59.92\% Micro Recall on CHAOS-Bench, surpassing GRPO and GRPO+OPD (fixed-weight) by 8.45 and 4.43 percentage points, respectively, while achieving an Overall score of 91.18 on OmniDocBench v1.6.
\end{abstract}

\section{Introduction}

Document parsing aims to faithfully convert the text and structure in page images into machine-readable representations. Vision-language models (VLMs) integrate text recognition, layout understanding, and structured generation, advancing complex document parsing~\citep{bai2025qwen25vl,poznanski2025olmocr2}. Yet strong parsing ability does not guarantee faithful transcription: misspellings, visually confusable characters, or unexpected words may be replaced with linguistically plausible expressions that contradict the image. This linguistic-prior hallucination turns reading into implicit rewriting~\citep{PAR,lee2026faithc4}. We focus on OCR transcription faithfulness: when visual evidence conflicts with linguistic priors, models should preserve the content actually visible in the image, even when the source itself contains errors, while maintaining reliable parsing of ordinary text and document structure.

\begingroup
\setlength{\intextsep}{5pt}
\begin{figure}[!t]
    \centering
    \setlength{\abovecaptionskip}{4pt}
    \includegraphics[width=\columnwidth]{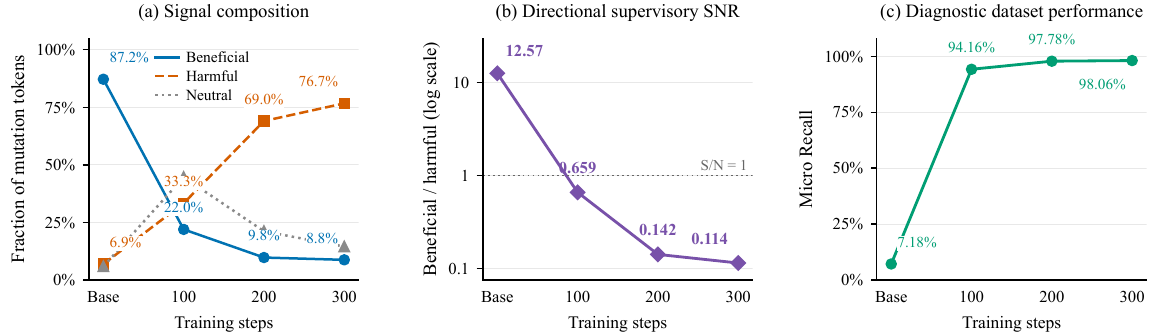}
    \par\nointerlineskip
    \caption{\textbf{Student performance improves while directional supervisory SNR declines.} The same frozen teacher scores GRPO checkpoints offline on a separate analysis set of 1,000 synthetic pages. (a) Signal composition on perturbed-word-associated tokens: relative to student probabilities at the same prefixes, teacher signals are beneficial when favoring correct emitted tokens or disfavoring errors, and potentially harmful in the reverse direction; small differences are neutral (Appendix~\ref{app:teacher_dynamics}). (b) Beneficial-to-harmful token-count ratio on a log scale; the dotted line marks equal counts. (c) Student perturbed-word Micro Recall.}
    \label{fig:teacher_supervision_rl}
\end{figure}
\endgroup

To optimize anomalous-content retention and transcription quality, we adopt Group Relative Policy Optimization~\citep[GRPO;][]{shao2024deepseekmath}, which has been applied to document parser post-training~\citep{poznanski2025olmocr2,wang2026infinityparser}. Unlike supervised fine-tuning (SFT), which learns under reference prefixes, GRPO uses relative rewards within groups of student-generated responses. However, identical group rewards yield zero group-relative advantages~\citep{yu2025dapo}, while rarely sampled faithful outputs receive limited direct reinforcement. We therefore complement GRPO with on-policy distillation (OPD), which provides token-level supervision at student-generated prefixes~\citep{agarwal2024opd}.

Recent work has explored regulating on-policy teacher supervision through student-aware target reformulation~\citep{jang2026veto} or trajectory-level selection~\citep{akhondzadeh2026rgopd}. Our focus is complementary: we study how the utility of a fixed teacher changes with the student's current task performance, and use this observation to regulate when and how strongly the teacher intervenes during joint GRPO post-training. Offline analysis of GRPO-only checkpoints shows rising perturbed-word Micro Recall (7.18\% to 98.06\%) and declining directional supervisory signal-to-noise ratio (SNR; 12.57 to 0.114) on a fixed analysis set (Figure~\ref{fig:teacher_supervision_rl}).

Motivated by these observations, we propose GAD-RL, a policy-state-aware framework that regulates teacher intervention according to the student's current task performance. Group-level gating determines when teacher supervision remains active, while reward-aware attenuation continuously reduces its strength as group performance improves. Separately, to avoid overly strong local FKL updates under substantial teacher--student disagreement, we use a simple student-probability weight to moderate the distillation gradient without altering the teacher target.

We evaluate GAD-RL on GlitchText~\citep{PAR} and CHAOS-Bench~\citep{li2026hunyuanocr15}, covering controlled text anomalies and character perturbations in realistic document layouts. On CHAOS-Bench, GAD-RL improves perturbed-word Micro Recall from 51.47\% for GRPO to 59.92\%, a gain of \textbf{8.45 percentage points}, while maintaining comparable general document-parsing performance (Table~\ref{tab:main_glitchtext}). Comparisons with fixed-weight distillation, step-based decay, and low-score response selection examine state-aware regulation against simpler controls (Section~\ref{sec:distillation_strength_selection}).

Our main contributions are:
\begin{itemize}
    \item \textbf{Revealing changes in teacher supervision with student state.} Through token-level analysis, we identify a decline in the same frozen teacher's directional supervisory signal-to-noise ratio as the student improves, providing empirical motivation for dynamically regulating teacher intervention.
    \item \textbf{Introducing state-aware teacher intervention for OCR faithfulness.} Based on the observed reward-dependent variation in supervision quality, GAD-RL combines group-level mastery gating with continuous reward-aware attenuation to regulate when and how strongly a frozen teacher intervenes during joint GRPO post-training. We additionally use student-probability-weighted FKL to moderate large local distillation updates under teacher--student disagreement.
    \item \textbf{Improving transcription faithfulness across student backbones.} GAD-RL improves CHAOS-Bench Recall over GRPO by 8.45 and 3.82 percentage points on Qwen3.5-2B and Qwen3-VL-2B, respectively, while maintaining comparable OmniDocBench performance. Ablations and controlled comparisons further support the effectiveness of regulating teacher intervention according to the student's current task performance.
\end{itemize}

\section{Related Work}
\label{sec:related_work}

\textbf{Document parsing.} LayoutLM and LayoutLMv3 integrate text, image, and layout features~\citep{huang2022layoutlmv3}; Donut, Pix2Struct, and Nougat recover structured content from images~\citep{kim2022donut,lee2023pix2struct,blecher2024nougat}. RL-based parsing uses verifiable rewards in olmOCR~2~\citep{poznanski2025olmocr2} and layout-aware rewards in Infinity-Parser~\citep{wang2026infinityparser}. OvisOCR2 combines real-document annotations with synthetic HTML-derived pages, then uses SFT, RL on a larger branch, OPD into a compact parser, and model fusion~\citep{lu2026ovisocr2}. OmniDocBench measures general parsing quality~\citep{ouyang2025omnidocbench}. However, current VLM-based parsers remain vulnerable to text perturbations: FaithC4 reveals unfaithful rewriting and error amplification on unperturbed content~\citep{lee2026faithc4}, while CHAOS-Bench, introduced with HunyuanOCR-1.5, reports low perturbed-word recall across evaluated OCR models~\citep{li2026hunyuanocr15}.

\textbf{Multimodal hallucination mitigation.} VCD, OPERA, and VISTA modify decoding or visual representations~\citep{leng2024vcd,huang2024opera,li2025vista}. PAR targets OCR overcorrection through positional perturbation and attention recycling at inference time, without additional training~\citep{PAR}. Training methods use factual feedback~\citep{sun2024llavarlhf}, preference optimization~\citep{zhao2023hadpo,yu2024rlhfv}, or phrase-level alignment~\citep{sarkar2025halva}. Our work focuses on OCR transcription faithfulness: mitigating overcorrection driven by linguistic priors while preserving the text actually present in document images.

\textbf{On-policy learning and distillation.} GKD supports on-policy distillation jointly with RL~\citep{agarwal2024opd}. KDRL combines GRPO and reverse KL, with linear coefficient decay and reward-guided response/group masks; group distillation stops when any response succeeds~\citep{xu2025kdrl}. OPSD uses privileged-context self-distillation~\citep{zhao2026opsd}; TGPO supplies teacher-preferred next tokens at student prefixes under large policy divergence~\citep{liu2026tgpo}. Veto constructs student-dependent geometric targets~\citep{jang2026veto}.

\textbf{Teacher supervision reliability and control.} RG-OPD filters trajectories by agreement between verifier advantages and teacher--student log-likelihood gaps~\citep{akhondzadeh2026rgopd}. RSTG (\emph{Distill Where You Fail}) combines all-incorrect zero-variance group selection, teacher-confidence weighting, token selection, and SFT on correct teacher trajectories~\citep{han2026distillwhereyoufail}. I-SDPO routes all-incorrect groups to privileged self-distillation and any-success groups to GRPO~\citep{zhang2026isdpo}. PACED favors intermediate student pass rates based on cross-problem gradient SNR~\citep{xu2026paced}. Analyzing teacher noise, \citet{ding2026opd} find that fixed negative advantages on low-probability sampled tokens can match OPD in their reverse-KL reasoning-distillation settings. We track a fixed teacher's correctness-aware beneficial-to-harmful token-count ratio across OCR student checkpoints as transcription improves. GAD-RL combines group gating and continuous group-mean-reward attenuation with GRPO; student-probability weighting separately moderates local FKL updates.

\section{Methods}
\label{sec:methods}

GAD-RL combines GRPO with state-aware on-policy distillation from a frozen teacher. Group-level gating and reward-aware attenuation weaken or disable teacher guidance as task performance improves, while student-adaptive forward KL scales token-level updates. GRPO remains active throughout training.

Let $(x,z)\sim\mathcal D$ be a document example, where $x$ contains the page image and parsing instruction, and $z$ is the ground truth (GT). The student is $\pi_\theta$. The frozen teacher $\pi_{\theta_{\mathrm T}}$ receives $x_{\mathrm T}$, a text prompt combining $z$ with a content-preserving rewrite instruction, without the page image. For each input, the rollout policy $\pi_{\theta_{\mathrm{old}}}$ generates a group $\mathcal Y=\{y^{(i)}\}_{i=1}^{G}$, with $y^{(i)}\sim\pi_{\theta_{\mathrm{old}}}(\cdot\mid x)$. Here, $i$ indexes responses, $t$ indexes tokens, and $T_i=|y^{(i)}|$ is the valid response length. Student and teacher score tokens under the same generated prefix $y_{<t}^{(i)}$. We write $\mathbb E_{\mathrm{roll}}$ for expectation over $(x,z)\sim\mathcal D$ and these sampled groups.

\paragraph{GRPO objective.}
Let $R_i=R(y^{(i)},z)$ be the task reward, defined in Section~\ref{sec:reward_function}, and let $\bar R$ and $\sigma_R$ be its group mean and standard deviation. The token-level policy ratio and sequence advantage are
\begin{equation}
    \rho_{i,t}(\theta)
    =\frac{\pi_\theta(y_t^{(i)}\mid x,y_{<t}^{(i)})}
    {\pi_{\theta_{\mathrm{old}}}(y_t^{(i)}\mid x,y_{<t}^{(i)})},
    \qquad A_i=\frac{R_i-\bar R}{\sigma_R},
    \qquad A_{i,t}=A_i.
\label{eq:grpo_ratio_advantage}
\end{equation}
When $\sigma_R=0$, we set all group advantages to zero. GRPO~\citep{shao2024deepseekmath} maximizes
\begin{equation}
\begin{aligned}
    \mathcal J_{\mathrm{GRPO}}(\theta)
    =\mathbb E_{\mathrm{roll}}\Biggl[&
    \frac{1}{G}\sum_{i=1}^{G}\frac{1}{T_i}\sum_{t=1}^{T_i}\\
    &\min\!\left(\rho_{i,t}(\theta)A_i,
    \operatorname{clip}\!\left(\rho_{i,t}(\theta),1-\varepsilon,1+\varepsilon\right)A_i\right)
    \Biggr],
\end{aligned}
\label{eq:grpo_objective}
\end{equation}
where $\varepsilon$ is the clipping parameter. Tokens are averaged within each response, and responses are averaged within each group.

\subsection{Mastery-Gated Distillation}
\label{sec:mastery_gating}

\textbf{When to distill.}
When the student can already generate a sequence that receives the maximum task reward for an input, that sequence requires no further correction under the current reward criterion. Continuing to match the teacher distribution may then introduce unnecessary supervision noise and conflict with task-reward optimization~\citep{han2026distillwhereyoufail}. We therefore disable group-level distillation when any sampled response reaches the success threshold. The gate is
\begin{equation}
    g=\mathbb I\!\left[\max_{i=1,\ldots,G}R_i<\tau\right].
    \label{eq:gad_group_gate}
\end{equation}
Here, $\mathbb I[\cdot]$ is the indicator function, and $\tau$ is chosen to tolerate minor formatting differences such as whitespace and heading levels. GRPO always uses all responses and reinforces those with above-average rewards when group rewards differ.

\subsection{Reward-Aware Attenuation}
\label{sec:reward_attenuation}

\textbf{How strongly to distill.}
A binary gate assigns the same weight to low-performing and nearly solved groups retained for distillation. Motivated by the lower directional supervisory SNR observed in higher-reward groups within individual checkpoints (Figure~\ref{fig:reward_binned_teacher_signals}), we use group-average task reward to reduce guidance as performance improves, while the gate determines whether distillation remains active.

\label{sec:reward_function}
For an output $y$ and its GT $z$, we combine normalized edit similarity $R_{\mathrm{edit}}$ with exact perturbed-word recall $R_{\mathrm{recall}}$:
\[
R(y,z)=
\begin{cases}
    R_{\mathrm{edit}}(y,z), & \text{regular documents},\\
    \eta R_{\mathrm{edit}}(y,z)+(1-\eta)R_{\mathrm{recall}}(y,z),
    & \text{text-perturbed documents}.
\end{cases}
\]
Here, $\eta\in[0,1]$ controls the balance between edit similarity and perturbed-word recall. Both components lie in $[0,1]$; their definitions are given in Appendix~\ref{app:gad_implementation}.

The group-mean reward is $\bar R=G^{-1}\sum_{i=1}^{G}R_i$. For groups with $g=1$, we use
\begin{equation}
    f_\kappa(\bar R)=\frac{e^{-\kappa\bar R}-e^{-\kappa}}{1-e^{-\kappa}}.
    \label{eq:gad_group_attenuation}
\end{equation}
The parameter $\kappa>0$ controls the decay rate. With $f_\kappa(0)=1$ and $f_\kappa(1)=0$, this factor continuously weakens teacher guidance as group-mean reward increases. Gating and attenuation together give the group weight $g f_\kappa(\bar R)$.

\subsection{Student-Adaptive Forward KL}
\label{sec:fkl_motivation}
\label{sec:student_weighted_fkl}

For the local gradient comparison at a fixed generated prefix, write $\mathbf p$ and $\mathbf q$ for the student and teacher probability vectors. Sampled-token estimators of reverse KL, $D_{\mathrm{KL}}(\mathbf p\|\mathbf q)$, naturally use student rollouts to provide on-policy token-level credit. Their direct signals act on sampled actions: they can penalize a student-preferred error without explicitly specifying which unsampled alternative to promote~\citep{liu2026tgpo,han2026distillwhereyoufail}. FKL instead supplies teacher-distribution targets at student-generated prefixes~\citep{agarwal2024opd}, including teacher-supported candidates absent from the sampled response.

This direct correction can be strong under teacher--student disagreement. Let $\mathbf h$ be the student logit vector, so $\mathbf p=\operatorname{softmax}(\mathbf h)$. Let $a$ denote the sampled token and $A$ its fixed group-relative advantage. The negative unclipped token-level GRPO surrogate, denoted $\ell_{\mathrm{GRPO}}$, at $\theta=\theta_{\mathrm{old}}$ and the full-vocabulary FKL loss $\ell_{\mathrm{FKL}}=D_{\mathrm{KL}}(\mathbf q\|\mathbf p)$ have logit gradients
\begin{equation}
\begin{aligned}
    \nabla_{\mathbf h}\ell_{\mathrm{GRPO}}&=A(\mathbf p-\mathbf e_a),\\
    \nabla_{\mathbf h}\ell_{\mathrm{FKL}}&=\mathbf p-\mathbf q,
\end{aligned}
\label{eq:grpo_fkl_gradient_comparison}
\end{equation}
where $\mathbf e_a$ is the one-hot vector for $a$. A teacher-supported candidate with very low student probability can receive a much stronger FKL signal than its policy-gradient signal, particularly when the student is highly confident in its sampled token. The FKL logit gradient is bounded, but its relative scale can be large. Since GRPO clipping does not constrain an added FKL term, excessive auxiliary contributions may destabilize policy updates.

We use Student-Adaptive Forward KL (SA-FKL) to moderate local updates when the student assigns low probability to the teacher's preferred candidate. Let $b_{i,t}$ denote the teacher's Top-1 token at the student-generated prefix. The probability weight and forward-KL term are
\begin{equation}
\begin{aligned}
    b_{i,t}&=\arg\max_{v\in\mathcal V}
    \pi_{\theta_{\mathrm T}}(v\mid x_{\mathrm T},y_{<t}^{(i)}),\\
    w_{i,t}(\theta)&=\operatorname{sg}\!\left[
    \pi_\theta(b_{i,t}\mid x,y_{<t}^{(i)})\right],\\
    d_{i,t}(\theta)&=D_{\mathrm{KL}}\!\left(
    \pi_{\theta_{\mathrm T}}(\cdot\mid x_{\mathrm T},y_{<t}^{(i)})
    \,\middle\|\,
    \pi_\theta(\cdot\mid x,y_{<t}^{(i)})\right).
\end{aligned}
\label{eq:gad_token_weight}
\end{equation}
Here, $\mathcal V$ is the vocabulary and $\operatorname{sg}$ stops gradient propagation through the weight. The token contribution is $w_{i,t}(\theta)d_{i,t}(\theta)$, so the student's probability of the teacher's Top-1 token scales the full local KL gradient without changing its direction. When the teacher favors a candidate with little student support, this weight attenuates the potentially strong FKL update. The gradient is derived in Appendix~\ref{app:adaptive_gradient_scale}.

\subsection{Joint optimization.}
We average the weighted KL terms over valid tokens in each group and apply the group controls inside the rollout expectation:
\begin{equation}
\begin{aligned}
    \mathcal J_{\mathrm{OPD}}(\theta)
    &=\mathbb E_{\mathrm{roll}}\!\left[
    g f_\kappa(\bar R)
    \frac{\sum_{i=1}^{G}\sum_{t=1}^{T_i}w_{i,t}(\theta)d_{i,t}(\theta)}
    {\sum_{i=1}^{G}T_i}\right],\\
    \mathcal J(\theta)
    &=\mathcal J_{\mathrm{GRPO}}(\theta)-\lambda\mathcal J_{\mathrm{OPD}}(\theta).
\end{aligned}
\label{eq:grpo-privileged-fkl}
\end{equation}
GAD-RL maximizes $\mathcal J$. Since $\mathcal J_{\mathrm{OPD}}$ is a distillation cost, it enters with a minus sign; $\lambda$ controls its strength. The gate and attenuation are computed separately for each sampled group and held fixed during the update. The vocabulary-truncated approximation is specified in Appendix~\ref{app:gad_implementation}.

\section{Experiments}
\label{sec:experiments}

\paragraph{Experimental setup.}
For each backbone, the teacher and student start from the same base model. The teacher is fine-tuned on 200,000 transcription-task examples. Document-parsing training uses a mixed dataset of 14,400 samples, with a 3:2 ratio of general to text-perturbed documents. General documents are sampled from Infinity-Doc2-5M~\citep{huang2026infinityparser2} and MonkeyDoc~\citep{li2025monkeyocr}. We cross-check their annotations against parsing outputs from PaddleOCR-VL-1.6~\citep{zhang2026paddleocrvl16} and discard samples with text similarity below 0.9. All student training strategies start from the base checkpoint and run for 300 steps with a learning rate of $10^{-6}$. Unless otherwise specified, we report benchmark results from the checkpoint after 300 training steps. Detailed training settings are provided in Appendix~\ref{app:gad_implementation}.

\paragraph{Baselines.}
We evaluate SFT, GRPO, and GRPO+OPD (fixed-weight) on both Qwen3-VL-2B and Qwen3.5-2B. \textbf{GRPO+OPD (fixed-weight)} adds a fixed-weight distillation loss directly to the GRPO objective on all responses. The default distillation coefficient is $\lambda=0.005$, used by both GAD-RL and GRPO+OPD (fixed-weight) in the main experiments.

\subsection{Benchmarks and Metrics}
\label{sec:eval_metrics}

\emph{GlitchText}~\citep{PAR} introduces controlled errors into familiar passages rendered on a plain background. We report Identification Rate (Ident) and Correction Rate (Cor), averaged equally across Chinese and English. \emph{CHAOS-Bench}~\citep{li2026hunyuanocr15} introduces character corruptions into 500 academic-paper page images with realistic layouts; we report Micro Recall. For general document parsing, we use \emph{OmniDocBench v1.6}~\citep{ouyang2025omnidocbench} and report its Overall score. Metric definitions, matching rules, and aggregation details are provided in Appendix~\ref{app:evaluation_metrics}.

\subsection{Main Results}
\label{sec:main_results}

Table~\ref{tab:main_glitchtext} compares the original model (Baseline), the post-training baselines, and GAD-RL for each student model.

\begingroup
\setlength{\intextsep}{6pt}
\begin{table}[!htb]
    \centering
    \setlength{\abovecaptionskip}{0pt}
    \setlength{\belowcaptionskip}{6pt}
    \caption{Transcription faithfulness and general document parsing, grouped by student model. Micro Recall is a ratio; other scores are percentages. Bold marks the best result within each model block.}
    \label{tab:main_glitchtext}
    \fontsize{8.5}{10.5}\selectfont
    \setlength{\tabcolsep}{6pt}
    \renewcommand{\arraystretch}{1.15}
    \begin{tabular*}{\linewidth}{@{\hspace{8pt}\extracolsep{\fill}}lcrrc@{\hspace{8pt}}}
        \toprule
        \multirow{2}{*}[-2pt]{\textbf{Method}}
        & \textbf{CHAOS-Bench}
        & \multicolumn{2}{c}{\textbf{GlitchText}}
        & \textbf{OmniDocBench v1.6} \\
        \cmidrule(lr){2-2}\cmidrule(lr){3-4}\cmidrule(lr){5-5}
        & Micro Recall$\uparrow$ & Ident$\uparrow$ & Cor$\downarrow$ & Overall$\uparrow$ \\
        \midrule
        \rowcolor[gray]{0.94}
        \multicolumn{5}{@{\hspace{8pt}}l@{\hspace{8pt}}}{\rule{0pt}{12pt}\textit{Student: Qwen3.5-2B}} \\
        Baseline   & 0.0402 & 76.48 & 14.94 & 80.06 \\
        SFT        & 0.3763 & 82.20 & 10.71 & 90.29 \\
        GRPO       & 0.5147 & 91.30 & 4.21 & 90.80 \\
        GRPO+OPD (fixed-weight) & 0.5549 & 91.31 & \textbf{3.14} & 90.90 \\
        \textbf{GAD-RL} & \textbf{0.5992} & \textbf{93.39} & 3.40 & \textbf{91.18} \\
        \midrule
        \rowcolor[gray]{0.94}
        \multicolumn{5}{@{\hspace{8pt}}l@{\hspace{8pt}}}{\rule{0pt}{12pt}\textit{Student: Qwen3-VL-2B}} \\
        Baseline   & 0.0235 & 83.91 & 8.28 & 46.39 \\
        SFT        & 0.3804 & 87.75 & 7.00 & 87.97 \\
        GRPO       & 0.5255 & 88.33 & 4.39 & 89.86 \\
        GRPO+OPD (fixed-weight) & 0.4167 & 89.26 & 4.69 & 87.67 \\
        \textbf{GAD-RL} & \textbf{0.5637} & \textbf{92.15} & \textbf{3.59} & \textbf{90.00} \\
        \bottomrule
    \end{tabular*}
\end{table}
\endgroup

On CHAOS-Bench, GAD-RL improves Micro Recall over GRPO by 8.45 and 3.82 percentage points on Qwen3.5-2B and Qwen3-VL-2B, respectively, and outperforms fixed-weight GRPO+OPD on both backbones.

On GlitchText, GAD-RL improves anomaly identification and reduces overcorrection relative to SFT and GRPO on both backbones; fixed-weight OPD retains a lower correction rate on Qwen3.5-2B. GAD-RL also maintains comparable general document-parsing performance on OmniDocBench.

\begin{figure}[!t]
    \centering
    \setlength{\abovecaptionskip}{4pt}
    \includegraphics[width=\linewidth]{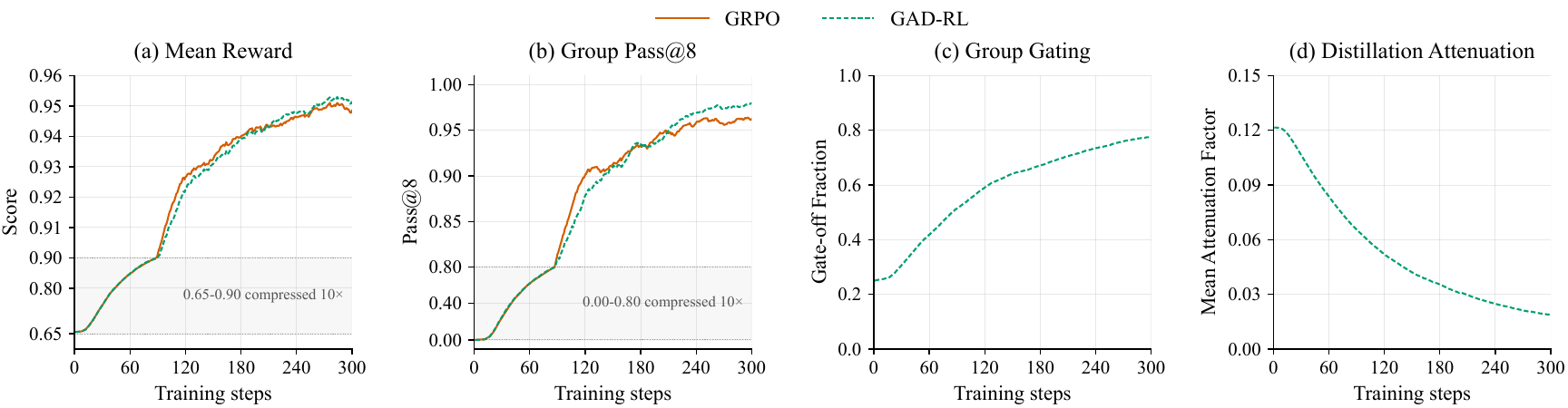}
    \caption{\textbf{GRPO and GAD-RL training dynamics.} (a) Mean reward. (b) Fraction of groups with at least one of eight responses achieving perturbed-word Recall of 1. For GAD-RL, (c) the fraction of groups with distillation disabled by the gate and (d) the mean attenuation factor among the remaining active groups.}
    \label{fig:grpo_gad_training_dynamics}
\end{figure}

\paragraph{Training dynamics.}
GRPO leads early in training, but GAD-RL overtakes it and finishes with higher mean reward and perturbed-word Pass@8 (Figure~\ref{fig:grpo_gad_training_dynamics}a,b).

\textbf{The two controls play complementary roles over training.} The gate disables an increasing fraction of groups, while the mean attenuation factor among active groups falls from about 0.12 to 0.02 (Figure~\ref{fig:grpo_gad_training_dynamics}c,d). Teacher guidance therefore weakens even on groups that remain active.

\Needspace{4\baselineskip}
\subsection{Ablation Studies}

\paragraph{KL formulation in on-policy distillation.}
\label{sec:kl_formulation_ablation}
Following prior work on divergence choices in on-policy distillation~\citep{agarwal2024opd}, we compare K1 PG, reverse KL (RKL), Jensen--Shannon divergence (JSD), FKL, and SA-FKL on Qwen3.5-2B within the GRPO+OPD (fixed-weight) setup, replacing only the distillation loss while keeping the coefficient at $\lambda$ and all other settings fixed.

\Needspace{17\baselineskip}
\begingroup
\setlength{\intextsep}{5pt}
\begin{wrapfigure}{r}{0.5\columnwidth}
    \centering
    \setlength{\abovecaptionskip}{4pt}
    \includegraphics[width=\linewidth]{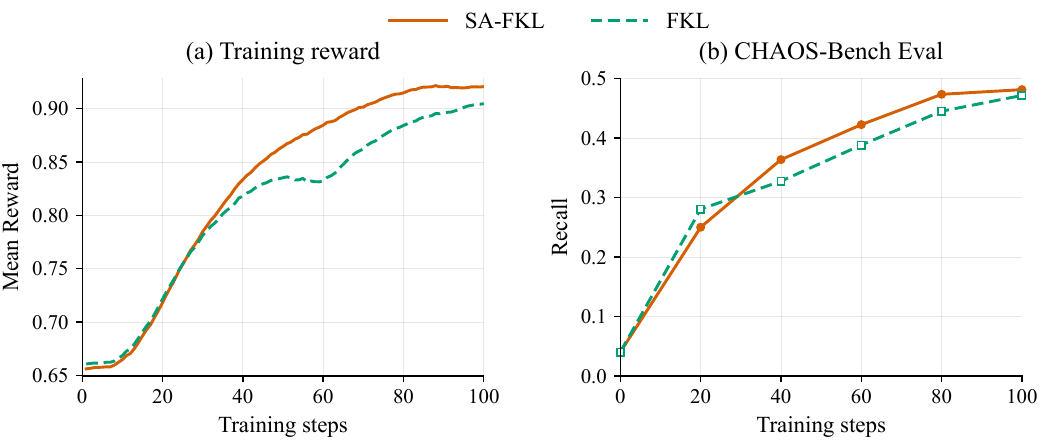}
    \caption{\textbf{FKL and SA-FKL training dynamics.} With otherwise identical settings: (a) mean training reward (EMA 0.95); (b) CHAOS-Bench Micro Recall, with a step-0 reference of 0.0402.}
    \label{fig:fkl_safkl_training_reward}
\end{wrapfigure}

SA-FKL achieves the highest CHAOS-Bench Recall after 100 steps (Table~\ref{tab:kl_formulation_ablation}). FKL learns faster initially but shows a reward dip around steps 50--60; SA-FKL is smoother over this interval and achieves higher Recall from step 40 onward (Figure~\ref{fig:fkl_safkl_training_reward}).

\paragraph{Distillation coefficient.}
Among the four nonzero coefficients in Table~\ref{tab:distillation_coefficients}, only $\lambda=0.005$ outperforms GRPO. Neither halving nor increasing this value helps, showing that fixed-weight OPD is sensitive to coefficient selection.

\par
\ifnum\value{WF@wrappedlines}>1
    \vspace{\dimexpr\value{WF@wrappedlines}\baselineskip-\baselineskip\relax}
\fi
\WFclear
\endgroup

\begin{table}[H]
    \centering
    \setlength{\abovecaptionskip}{0pt}
    \setlength{\belowcaptionskip}{6pt}
    \setlength{\tabcolsep}{12pt}
    \renewcommand{\arraystretch}{1.15}
    \sisetup{mode=text,detect-weight=true,table-format=1.4,table-column-width=4.4em}
    \begin{minipage}[t]{0.43\linewidth}
        \centering
        \vspace{0pt}
        \caption[Distillation objectives on CHAOS-Bench.]{\raggedright Distillation objectives on \mbox{CHAOS-Bench} at step 100.\strut}
        \label{tab:kl_formulation_ablation}
        \small
        \begin{tabular}{@{\hspace{8pt}}>{\raggedright\arraybackslash}p{8em} S@{\hspace{8pt}}}
            \toprule
            \textbf{Objective} & {\textbf{Recall}$\uparrow$} \\
            \midrule
            K1 PG & 0.4500 \\
            RKL & 0.4441 \\
            JSD & 0.4578 \\
            FKL & 0.4716 \\
            \textbf{SA-FKL} & \bfseries 0.4814 \\
            \bottomrule
        \end{tabular}
    \end{minipage}\hspace{0.06\linewidth}%
    \begin{minipage}[t]{0.43\linewidth}
        \centering
        \vspace{0pt}
        \caption[Fixed-weight distillation coefficients.]{\raggedright Fixed-weight distillation at different coefficients.\strut}
        \label{tab:distillation_coefficients}
        \small
        \begin{tabular}{@{\hspace{8pt}}>{\raggedright\arraybackslash}p{8em} S@{\hspace{8pt}}}
            \toprule
            \textbf{Coefficient} & {\textbf{Recall}$\uparrow$} \\
            \midrule
            $0$ (GRPO) & 0.5147 \\
            $0.5\lambda$ & 0.5010 \\
            $\lambda$ & \bfseries 0.5549 \\
            $5\lambda$ & 0.4873 \\
            $10\lambda$ & 0.4637 \\
            \bottomrule
        \end{tabular}
    \end{minipage}
\end{table}

\paragraph{Gating and attenuation.}
Table~\ref{tab:gad_controls_ablation} removes student-probability weighting (SW), mastery-gated distillation (MGD), or reward-aware attenuation (RAA) at $\lambda$. Removing SW or RAA causes larger Recall losses than removing MGD. The smaller incremental benefit of gating is consistent with RAA already suppressing high-reward groups: at $\bar R=0.9$, it retains only $1.83\%$ of the base coefficient before token weighting.

\begingroup
\setlength{\intextsep}{6pt}
\begin{table}[H]
    \centering
    \setlength{\abovecaptionskip}{0pt}
    \setlength{\belowcaptionskip}{6pt}
    \setlength{\tabcolsep}{12pt}
    \renewcommand{\arraystretch}{1.15}
    \sisetup{mode=text,detect-weight=true,table-format=1.4,table-column-width=4.4em}
    \caption{Component ablations on CHAOS-Bench at $\lambda$.}
    \label{tab:gad_controls_ablation}
    \label{tab:distillation_ablation}
    \small
    \begin{tabular}{@{\hspace{8pt}}>{\raggedright\arraybackslash}p{8em} S@{\hspace{8pt}}}
        \toprule
        \textbf{Variant} & {\textbf{Recall}$\uparrow$} \\
        \midrule
        GRPO & 0.5147 \\
        \textbf{GAD-RL (full)} & \bfseries 0.5992 \\
        \quad w/o SW & 0.5598 \\
        \quad w/o MGD & 0.5931 \\
        \quad w/o RAA & 0.5414 \\
        \bottomrule
    \end{tabular}
\end{table}
\endgroup

\section{Analysis}
\label{sec:analysis}

Building on Figure~\ref{fig:teacher_supervision_rl}, we examine whether the teacher prefers incorrect candidates at correct student outputs and how supervision quality varies with group reward within each checkpoint. We then compare teacher-intervention strategies through training experiments.

\subsection{Teacher Supervision as the Student Improves}
\label{sec:teacher_supervision_rl}

Lowering a correct token's probability does not necessarily imply an incorrect teacher Top-1 prediction. We therefore inspect teacher Top-1 predictions at correct student outputs, reusing Figure~\ref{fig:teacher_supervision_rl}'s 1,000 pages excluded from training, GRPO-only student checkpoints, and frozen teacher.

At the same student-generated prefix, we count a teacher error when its Top-1 token conflicts with the GT continuation. Among correctly transcribed perturbed-word-associated tokens, this rate rises from 9.52\% at step 100 to 42.88\% at step 300 (Table~\ref{tab:teacher_errors_correct_outputs}), showing an increasing preference for incorrect candidates at positions the student already transcribes correctly.

\begin{table}[!htbp]
    \centering
    \setlength{\abovecaptionskip}{0pt}
    \setlength{\belowcaptionskip}{5pt}
    \caption{Teacher Top-1 errors at correct student outputs. Counts refer to perturbed-word-associated tokens. The final columns give teacher errors as percentages of all eligible tokens and of correct student tokens.}
    \label{tab:teacher_errors_correct_outputs}
    \small
    \setlength{\tabcolsep}{7pt}
    \renewcommand{\arraystretch}{1.15}
    \begin{tabular}{@{\hspace{8pt}}lrrrrr@{\hspace{8pt}}}
        \toprule
        \textbf{Checkpoint} & \textbf{All tokens} & \textbf{Correct tokens} & \textbf{Teacher errors} & \textbf{\% of all} & \textbf{\% of correct} \\
        \midrule
        Base & 4,281 & 1,029 & 131 & 3.06 & 12.73 \\
        Step 100 & 10,325 & 10,159 & 967 & 9.37 & 9.52 \\
        Step 200 & 12,903 & 12,865 & 4,234 & 32.81 & 32.91 \\
        Step 300 & 13,391 & 13,366 & 5,731 & 42.80 & 42.88 \\
        \bottomrule
    \end{tabular}
\end{table}

Token counts vary across checkpoints because the generated responses differ, resulting in different numbers of tokens aligned to perturbed-word spans; detailed statistics and matching rules are provided in Appendix~\ref{app:teacher_dynamics}. 

\begin{figure}[!htbp]
    \centering
    \setlength{\abovecaptionskip}{4pt}
    \includegraphics[width=\linewidth]{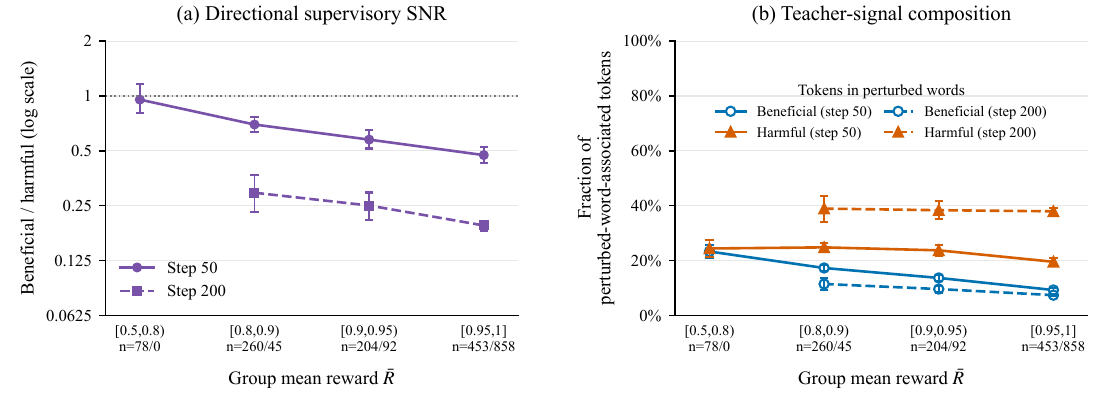}
    \caption{\textbf{Teacher supervision becomes less favorable as group reward increases.} GRPO+OPD groups at steps 50 and 200 are binned by mean reward $\bar R$. For perturbed-word-associated tokens: (a) beneficial-to-harmful count ratio (log scale; dotted line: equal counts); (b) beneficial and harmful fractions, with the remainder neutral. The absolute probability-change margin is 0.1. $n$ gives group counts at step 50 / 200; error bars show 95\% input-group-bootstrap confidence intervals.}
    \label{fig:reward_binned_teacher_signals}
\end{figure}

\paragraph{Reward dependence within each checkpoint.}
Across-checkpoint trends combine changes in training stage and task performance. We therefore examine reward dependence within GRPO+OPD checkpoints at steps 50 and 200, scoring current-policy rollouts with the same frozen teacher. Figure~\ref{fig:reward_binned_teacher_signals} partitions 995 matched diagnostic page groups by mean task reward; scoring and bootstrap details are in Appendix~\ref{app:reward_stratified_signals}.

Within both checkpoints, beneficial signals become less frequent as group reward increases, while harmful signals do not decline proportionally. Consequently, useful teacher corrections become scarcer relative to conflicting supervision as the student performs better on an input. This within-checkpoint pattern supports using group reward to regulate teacher intervention: a training-step schedule assigns the same coefficient to groups with different relative corrective opportunities. Reward-aware attenuation can instead reduce guidance on groups that already achieve high task reward.

\subsection{Distillation Control Strategies}
\label{sec:distillation_strength_selection}

\textbf{How should teacher intervention be controlled?} Section~\ref{sec:teacher_supervision_rl} motivates comparing uniform changes in strength, response selection, and decay over training steps with state-aware regulation. Building on the coefficient ablation in Table~\ref{tab:distillation_coefficients}, Table~\ref{tab:supervision_strength_selection} compares filtering and decay strategies. These Qwen3.5-2B comparisons retain the GRPO task loss and use a base coefficient of $\lambda=0.005$. All distillation-control comparisons use SA-FKL; fixed-weight OPD applies it to all responses. Recall is evaluated on CHAOS-Bench. Low-score OPD selects individual responses with total reward $R_i<0.5$; GAD-RL gates whole groups and attenuates their weight using group-mean reward.

\begin{table}[!htbp]
    \centering
    \setlength{\abovecaptionskip}{0pt}
    \setlength{\belowcaptionskip}{5pt}
    \renewcommand{\arraystretch}{1.15}
    \sisetup{mode=text,detect-weight=true,table-format=1.4,table-column-width=4.4em}

    \begin{minipage}[t]{0.76\linewidth}
        \centering
        \vspace{0pt}
        \caption[Distillation decay and filtering strategies.]{Distillation decay and filtering strategies. The best Recall is bold.}
        \label{tab:supervision_strength_selection}
        \small
        \setlength{\tabcolsep}{5pt}
        \begin{tabular}{@{\hspace{8pt}}lc S@{\hspace{8pt}}}
            \toprule
            \textbf{Method} & \textbf{Coefficient}$^{1}$ & {\textbf{Recall}$\uparrow$} \\
            \midrule
            GRPO & $0$ & 0.5147 \\
            \quad + Linear-decay OPD & $\lambda\,s(u)$ & 0.5693 \\
            \quad + Low-score OPD & $\lambda$ & 0.5775 \\
            \midrule
            \multirow{2}{*}{\textbf{GAD-RL}} & $\lambda g f_\kappa(\bar R)$ & \bfseries 0.5992 \\
            & $10\lambda g f_\kappa(\bar R)$ & 0.5882 \\
            \bottomrule
        \end{tabular}
    \end{minipage}
    \par\smallskip
    \begin{minipage}{0.76\linewidth}
        \footnotesize
        $^{1}$The base coefficient is $\lambda=0.005$; token-level weights are omitted. The gate $g$ and attenuation $f_\kappa(\bar R)$ are defined in Sections~\ref{sec:mastery_gating} and~\ref{sec:reward_attenuation}, respectively. The factor $s(u)$ decreases linearly with training step $u$.
    \end{minipage}
\end{table}

\paragraph{Response selection and temporal decay.}
Compared with GRPO+OPD (fixed-weight), low-score OPD and linear-decay OPD improve Recall by 2.26 and 1.44 percentage points, respectively. These gains are consistent with the changing utility of teacher supervision: response selection concentrates guidance on outputs needing correction, while temporal decay relaxes teacher constraints as training progresses.

\paragraph{GAD-RL coefficient.}
Increasing GAD-RL's base coefficient tenfold lowers Recall by 1.10 percentage points; both tested settings outperform GRPO. The strong attenuation on active groups shown in Figure~\ref{fig:grpo_gad_training_dynamics}d may help explain this modest change: teacher guidance weakens as group rewards improve, even when the base coefficient is larger.

\section{Conclusions}

We presented GAD-RL, which combines GRPO with group-gated, reward-attenuated, and student-weighted on-policy distillation. Offline analysis shows that teacher--student disagreement increasingly affects correctly transcribed perturbed-word-associated tokens as the student improves, while favorable suppression of residual errors remains common. GAD-RL improves transcription faithfulness over SFT and GRPO on two student models while maintaining comparable OmniDocBench performance. On Qwen3.5-2B, GAD-RL outperforms GRPO at both tested distillation coefficients. These findings support adapting teacher intervention to the student's evolving task performance and local distributions.

\paragraph{Limitations and future work.}
Our evaluation is limited to OCR transcription faithfulness in document parsing. Future work will explore GAD-RL on other tasks to assess whether policy-state-aware distillation control generalizes beyond OCR.

\newpage
\subsection*{AI use statement}

In this work, we used generative AI tools to assist with literature search and summarization, the design of mathematical notation, and formula verification. AI tools also generated most of the code used for synthetic data generation, figure visualization, and method implementation. The authors manually reviewed and validated the code and take responsibility for the final manuscript, mathematical claims, implementation, and reported results.

\subsection*{Reproducibility statement}

Appendix~\ref{app:data_synthesis} describes data synthesis, Appendix~\ref{app:token_visualization} details teacher-signal analysis, and Appendix~\ref{app:gradient_analysis} provides the gradient derivations. Training configurations and objectives are specified in Appendix~\ref{app:gad_implementation}, and evaluation metrics in Appendix~\ref{app:evaluation_metrics}. Upon acceptance, we will release the training, evaluation, and analysis code, the data-synthesis pipeline, teacher-training data, and the 1,000-page analysis set. Data will be distributed as files where their licenses permit, or through source manifests and reconstruction instructions where redistribution is restricted.

\subsection*{Ethics statement}

This work studies faithful transcription using synthetic text perturbations and existing document benchmarks. Synthetic documents are derived from arXiv LaTeX sources, whose reuse conditions depend on each source's license. Third-party resources, including GlitchText, CHAOS-Bench, and OmniDocBench, remain subject to their original licenses and terms. Releases of derived data will preserve source attribution and respect redistribution restrictions. Perturbed text is intended to test transcription fidelity and should not be treated as factual content. Improved transcription faithfulness does not establish the truth of the underlying documents.

\newpage
\bibliography{latex/iclr2027_conference}
\bibliographystyle{latex/iclr2027_conference}

\newpage
\appendix

\section{Data Synthesis}
\label{app:data_synthesis}

\subsection{Text-perturbed documents}
\label{app:multimodal_synthesis}

We construct text-perturbed documents from arXiv LaTeX projects. Text, headings, formulas, and tables are reflowed into academic layouts to produce page images and matching Markdown targets. The target preserves the anomalous text printed in the image.

\begingroup
\setlength{\intextsep}{4pt}
\begin{wrapfigure}{r}{0.5\columnwidth}
    \centering
    \setlength{\abovecaptionskip}{4pt}
    \includegraphics[width=\linewidth]{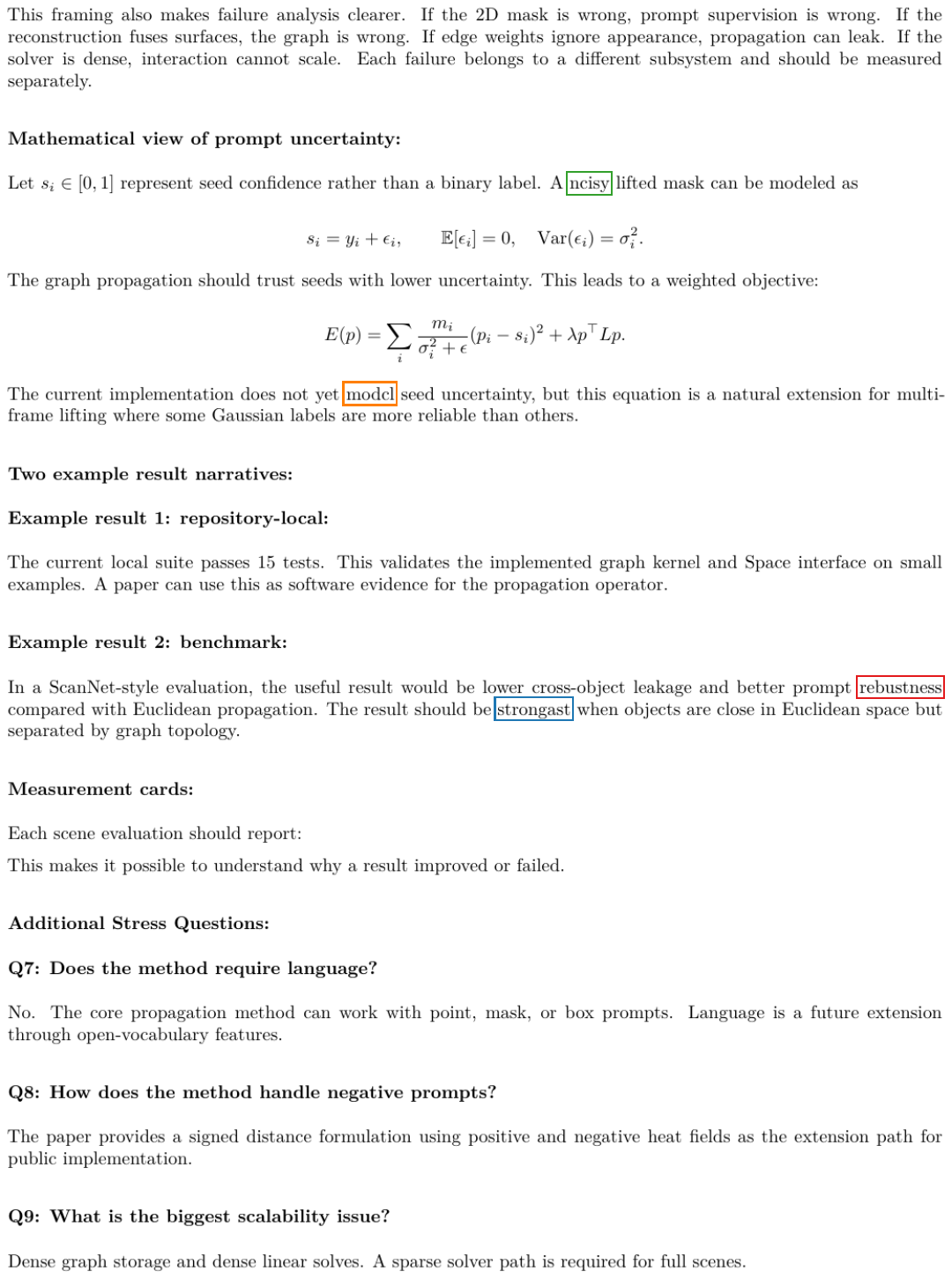}
    \caption{Example of a synthetically perturbed document page. Boxes highlight character-level substitutions in ``noisy'', ``model'', ``robustness'', and ``strongest''.}
    \label{fig:synthetic_page_example}
\end{wrapfigure}

\paragraph{Page construction.}
Source extraction removes citations, cross-references, bibliography entries, and figure content. Tables and formulas retain their source-derived content; tables use HTML in the target. Pages preserve the association of tables with captions and headings with following content. Unsupported structures and overflowing pages are rejected.

\paragraph{Controlled perturbations.}
Candidate words contain at least four letters, occur uniquely on the page, and have unambiguous source positions. Each edit replaces one lowercase letter without changing word length, using a shared weighted character-pair table. We request three edits per page with probability 0.4 and four with probability 0.6. Headings, captions, tables, formulas, code, and numerical content are protected; replacements colliding with existing words are skipped. Figure~\ref{fig:synthetic_page_example} shows an example.

\paragraph{Verification.}
Substitutions are applied consistently to the rendering source and target. Extracted PDF text and positions verify consistency and detect overflow, without generating or repairing the target. A separate analysis set contains 1,000 pages from 147 papers, including 140 table pages and 3,564 perturbed words.
\par
\ifnum\value{WF@wrappedlines}>1
    \vspace{\dimexpr\value{WF@wrappedlines}\baselineskip-\baselineskip\relax}
\fi
\WFclear
\endgroup

\subsection{Teacher rewrite data}
\label{app:teacher_synthesis}

The teacher is fine-tuned on 200,000 transcription-task examples. We build copying and controlled-format rewrite pairs from source-derived transcriptions, without OCR or LLM content rewriting. Samples contain 1,000--7,800 tokens with tables kept intact, and train/validation splits are made by source paper. Approximately 10\% of eligible word occurrences are perturbed, with at least three edits, using the same character-pair table.

Input and target preserve identical content, including anomalous words, numbers, formulas, and tables. Only existing heading prefixes and an optional outer Markdown fence may change; heading-level changes need not preserve hierarchy. Validation requires exact target reconstruction using only these permitted edits.

\section{Token Classification and Teacher-Signal Analysis}
\label{app:token_visualization}

\subsection{Fixed-trajectory scoring}

For each page, we generate a response using the Qwen3.5-2B-based student with thinking disabled and a maximum generation length of 8,192 tokens. Student and teacher score the same emitted token ID at the same response prefix: the student is conditioned on the image, and the teacher on a heading-rewrite prompt containing the reference text. All probabilities are computed over the full vocabulary.

\subsection{Token alignment, categories, and correctness}

Decoded responses are aligned to the reference after Unicode NFKC normalization and removal of recognized markup and whitespace. Annotated perturbed words must match unambiguous whole-word reference spans. Ambiguous or invalid alignments are excluded from correctness statistics.

Each response token is assigned one category, in priority order: \emph{perturbed-word-associated} if it overlaps the aligned full span of a perturbed word; \emph{formatting} if it contains only whitespace or markup; and \emph{body} otherwise. Perturbed-word association includes correct copies and overcorrections, not just the edited character. Visible numbers, punctuation, formulas, table-cell text, and code remain content; tokens mixing markup and content are body unless perturbed-word overlap takes priority.

Categories do not imply correctness. A content token is correct only when all its aligned normalized content characters match the reference; a correct subtoken within an erroneous word can therefore remain correct. Statistics count eligible response tokens equally, excluding prompts, trailing special tokens, and invalid alignments.

\subsection{Teacher information reliability}
\label{app:teacher_reliability}
\label{sec:teacher_reliability}

We compare the original Qwen3.5-2B self-teacher and the SFT teacher specified in Appendix~\ref{app:gad_implementation} on the same 1,000 student responses. Both receive identical reference text and the training-time teacher prompt. Correctness denominators exclude 5,179 records with inconsistent decoded-token offsets.

\begingroup
\setlength{\intextsep}{4pt}
\begin{wrapfigure}{r}{0.5\columnwidth}
    \centering
    \setlength{\abovecaptionskip}{4pt}
    \includegraphics[width=\linewidth]{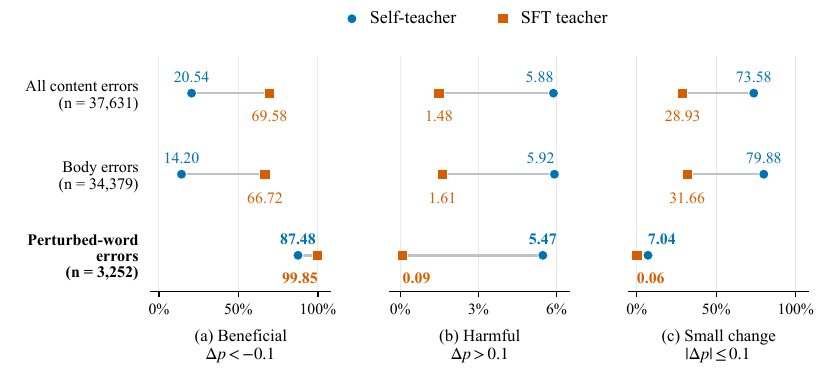}
    \caption{Teacher signals on erroneous tokens. For the same emitted token, $\Delta p<-0.1$ is beneficial suppression and $\Delta p>0.1$ is harmful reinforcement. Panel (b) uses an enlarged horizontal scale.}
    \label{fig:privileged_error_token_signals}
\end{wrapfigure}

\paragraph{Error suppression.}
With $\Delta p=p^{\mathrm T}-p^{\mathrm S}$, the SFT teacher suppresses 69.58\% of the 37,631 erroneous content tokens by more than 0.1, compared with 20.54\% for the self-teacher. Harmful reinforcement falls from 5.88\% to 1.48\% (Figure~\ref{fig:privileged_error_token_signals}). On perturbed-word errors, beneficial suppression reaches 99.85\%. These directions concern the emitted token; suppression alone does not verify a faithful alternative.

\paragraph{Teacher-choice ablation.}
Table~\ref{tab:teacher_training_ablation} uses the same Qwen3.5-2B GRPO and default SFT-teacher results as Table~\ref{tab:main_glitchtext}. The SFT teacher improves Recall over the self-teacher by 5.51 percentage points, supporting the use of a teacher trained for transcription.
\par
\ifnum\value{WF@wrappedlines}>1
    \vspace{\dimexpr\value{WF@wrappedlines}\baselineskip-\baselineskip\relax}
\fi
\WFclear
\endgroup

\begin{table}[!htb]
    \centering
    \setlength{\belowcaptionskip}{5pt}
    \caption{Teacher-choice ablation for GAD-RL on CHAOS-Bench.}
    \label{tab:teacher_training_ablation}
    \small
    \setlength{\tabcolsep}{12pt}
    \renewcommand{\arraystretch}{1.15}
    \sisetup{mode=text,detect-weight=true,table-format=1.4,table-column-width=4.4em}
    \begin{tabular}{@{\hspace{8pt}}ll S@{\hspace{8pt}}}
        \toprule
        \textbf{Method} & \textbf{Teacher} & {\textbf{Recall}$\uparrow$} \\
        \midrule
        GRPO & None & 0.5147 \\
        GAD-RL & Self-teacher (2B) & 0.5441 \\
        GAD-RL & SFT Qwen3.5-2B & 0.5992 \\
        \bottomrule
    \end{tabular}
\end{table}

\subsection{Frozen-teacher supervision across GRPO checkpoints}
\label{app:teacher_dynamics}

Figure~\ref{fig:teacher_supervision_rl} evaluates each GRPO-only checkpoint on the same analysis set described in Section~\ref{sec:teacher_supervision_rl}, with teacher scoring performed offline. Initialization, training data, and the teacher prompt match the main experiments (Appendix~\ref{app:gad_implementation}). Base is the RL initial checkpoint (step 0); students receive no OPD updates.

\paragraph{Signal definition.}
For an eligible perturbed-word-associated token, let $c\in\{0,1\}$ indicate correctness, define $\Delta p=p_{\mathrm T}-p_{\mathrm S}$, and set $h=(2c-1)\Delta p$. A signal is beneficial for $h>0.1$, harmful for $h<-0.1$, and neutral otherwise. Invalid alignments and formatting-only tokens are excluded. With counts $B$, $H$, and $U$, the plotted fractions use denominator $N=B+H+U$, and directional SNR is $B/H$.

\begin{table}[!htb]
    \centering
    \setlength{\belowcaptionskip}{4pt}
    \caption{Fixed-teacher signal counts and perturbed-word Recall on the analysis set. Signal statistics use eligible emitted tokens; Recall uses all 3,564 annotated words, including omissions.}
    \label{tab:teacher_dynamics_counts}
    \small
    \setlength{\tabcolsep}{8pt}
    \renewcommand{\arraystretch}{1.15}
    \begin{tabular}{@{\hspace{8pt}}lrrrrrr@{\hspace{8pt}}}
        \toprule
        \textbf{Student} & $N$ & \textbf{Beneficial} & \textbf{Harmful} & \textbf{Neutral} & $B/H$ & \textbf{Recall (\%)} \\
        \midrule
        Base & 4,281 & 3,732 & 297 & 252 & 12.566 & 7.18 \\
        Step 100 & 10,325 & 2,270 & 3,443 & 4,612 & 0.659 & 94.16 \\
        Step 200 & 12,903 & 1,261 & 8,908 & 2,734 & 0.142 & 97.78 \\
        Step 300 & 13,391 & 1,175 & 10,273 & 1,943 & 0.114 & 98.06 \\
        \bottomrule
    \end{tabular}
\end{table}

\paragraph{Token-count variation.}
These counts follow the tokens each checkpoint generates. Across the same 3,493 perturbed words represented at every checkpoint, the average number of tokens per word rises from 1.22 at Base to 2.94, 3.65, and 3.79 at steps 100, 200, and 300. Finer output tokenization accounts for most of the increase, while the 1,000 input pages remain fixed.

\paragraph{Preservation as the student improves.}
Recall uses normalized reference-span matching. From Base through steps 100, 200, and 300, the teacher assigns probabilities more than 0.1 below the student's to 28.57/33.88/69.24/76.86\% of correctly transcribed perturbed-word-associated tokens. Favorable suppression signals on remaining erroneous tokens occur at rates of 99.85/97.59/100/100\%, with denominators 3,252/166/38/25. At step 300 all potentially harmful signals occur on correct tokens. The aggregate decline thus accompanies greater directional disagreement on currently correct output, while favorable signals on residual errors remain prevalent.

\paragraph{Teacher Top-1 at suppressed correct tokens.}
We split correct perturbed-word-associated tokens with $\Delta p<-0.1$ by teacher Top-1 compatibility (Table~\ref{tab:teacher_top1_drop}). Using Appendix~\ref{app:token_visualization}'s normalization, we require consecutive exact-match GT anchors for the emitted token. Nonempty Top-1 content matching the GT continuation's prefix is compatible, including alternative tokenizations; nonmatching content is incompatible. Unavailable alignments and formatting-only or undecodable candidates remain unresolved. This tests local compatibility, not whole-word correctness.

\begin{table}[!htb]
    \centering
    \setlength{\belowcaptionskip}{4pt}
    \caption{Teacher Top-1 at correct perturbed-word-associated tokens with $\Delta p<-0.1$. Entries are counts (percent of $N$), including unresolved cases in the denominator.}
    \label{tab:teacher_top1_drop}
    \small
    \setlength{\tabcolsep}{8pt}
    \renewcommand{\arraystretch}{1.15}
    \begin{tabular}{@{\hspace{8pt}}lrrrr@{\hspace{8pt}}}
        \toprule
        \textbf{Student} & $N$ & \textbf{GT-compatible} & \textbf{Incompatible} & \textbf{Unresolved} \\
        \midrule
        Base & 294 & 136 (46.26) & 131 (44.56) & 27 (9.18) \\
        Step 100 & 3,442 & 2,413 (70.10) & 967 (28.09) & 62 (1.80) \\
        Step 200 & 8,908 & 4,615 (51.81) & 4,234 (47.53) & 59 (0.66) \\
        Step 300 & 10,273 & 4,458 (43.40) & 5,731 (55.79) & 84 (0.82) \\
        \bottomrule
    \end{tabular}
\end{table}

These diagnostics count tokens on the analysis set. Checkpoints emit different token sets, and small remaining error subsets limit rate comparisons.

\subsection{Reward-stratified teacher supervision}
\label{app:reward_stratified_signals}

Figure~\ref{fig:reward_binned_teacher_signals} uses Qwen3.5-2B GRPO+OPD checkpoints at steps 50 and 200 and the same frozen SFT teacher. The two checkpoints share 995 matched diagnostic page groups. Of the original 1,000 diagnostic pages, five are excluded because model inference repeatedly failed for these inputs despite retries; 

We use the correctness-aware signal definition in Appendix~\ref{app:teacher_dynamics} with an absolute probability-change margin of 0.1. Within each bin, beneficial and harmful fractions divide the respective pooled token counts by all evaluable perturbed-word-associated tokens, including neutral tokens. Directional SNR is the ratio of the pooled beneficial and harmful counts. Error bars are 95\% bootstrap confidence intervals from 2,500 resamples of input groups within each reward bin, keeping responses from the same input clustered. Empty bins have no estimate.

\section{Gradient Scale of Controlled Distillation}
\label{app:gradient_analysis}

\subsection{Policy gradients and unweighted FKL}
\label{app:relative_gradient_scale}

All gradients below are with respect to the student logits $\mathbf h$ at unit temperature. Using Equation~\ref{eq:grpo_fkl_gradient_comparison}, let $p(a)=1-\xi$ and $q(a)\leq1-\delta$ for $0<\xi<\delta\leq1$: $\xi$ is the student's probability mass away from sampled token $a$, and $\delta$ lower-bounds the corresponding teacher mass. At $\theta=\theta_{\mathrm{old}}$, before clipping and group controls,
\begin{equation}
 \|\nabla\ell_{\mathrm{GRPO}}\|_1=2|A|\xi,
 \qquad
 2(\delta-\xi)\leq\|\nabla\ell_{\mathrm{FKL}}\|_1\leq2.
\label{eq:pg_fkl_scale_bounds}
\end{equation}
For fixed disagreement and bounded nonzero advantages, FKL can be much stronger than PG at a high-confidence error, although its logit gradient remains bounded. This comparison holds at fixed prefixes and does not differentiate through the rollout distribution.

\subsection{Teacher-preferred-token probability weighting}
\label{app:adaptive_gradient_scale}

At a fixed prefix, let $b=\arg\max_vq(v)$ be the teacher's Top-1 token, $w=\operatorname{sg}[p(b)]$, and $d=D_{\mathrm{KL}}(\mathbf q\|\mathbf p)$. The weighted token loss is $\ell_{\mathrm{SA\text{-}FKL}}=wd$. With the probability weight detached,
\begin{equation}
    \nabla_{\mathbf h}\ell_{\mathrm{SA\text{-}FKL}}
    =p(b)(\mathbf p-\mathbf q).
\label{eq:safkl_local_gradient}
\end{equation}
This scales the FKL gradient without changing its direction. If the teacher prefers a candidate to which the student assigns little probability, the local update is attenuated even when the student is highly confident in its sampled token $a$. In the high-confidence setting of Appendix~\ref{app:relative_gradient_scale}, $b\ne a$ implies $p(b)\leq\xi$, hence
\begin{equation}
    \|\nabla_{\mathbf h}\ell_{\mathrm{SA\text{-}FKL}}\|_1
    \leq2p(b)\leq2\xi.
\label{eq:safkl_high_confidence_bound}
\end{equation}
This trades correction strength for update moderation when the student assigns low probability to the teacher's preferred candidate.

For the Top-$K$ approximation, let $\mathbf q^{\mathcal K}$ retain the original teacher probabilities on $\mathcal K$ and be zero elsewhere, and set $s=\sum_{v\in\mathcal K}q(v)$. Writing $d^{\mathcal K}$ for the truncated KL term gives
\begin{equation}
\begin{aligned}
    \nabla_{\mathbf h}\!\left(wd^{\mathcal K}\right)
    &=p(b)(s\mathbf p-\mathbf q^{\mathcal K}),\\
    \left\|\nabla_{\mathbf h}\!\left(wd^{\mathcal K}\right)\right\|_1
    &\leq2sp(b).
\end{aligned}
\label{eq:topk_safkl_gradient}
\end{equation}
The scalar gate $g$, attenuation $f_\kappa(\bar R)$, and coefficient $\lambda$ further scale these local contributions. Actual parameter updates also depend on model Jacobians, batch aggregation, and the optimizer.

\section{Training Configuration and Objectives}
\label{app:gad_implementation}

\paragraph{Objective averaging.}
Equation~\ref{eq:grpo_objective} first averages valid tokens within each response, then averages responses within a group. The OPD term in Equation~\ref{eq:grpo-privileged-fkl} instead divides the sum of weighted token KL terms by $\sum_i T_i$ for each group, before taking the rollout expectation. Groups with $g=0$ contribute zero to OPD and remain in the GRPO objective. Rewards, advantages, and group controls are held fixed during policy updates. The equivalent minimized training loss is $-\mathcal J_{\mathrm{GRPO}}+\lambda\mathcal J_{\mathrm{OPD}}$.

\begin{table}[!htb]
    \centering
    \setlength{\belowcaptionskip}{4pt}
    \caption{Shared GAD-RL training configuration for Qwen3.5-2B and Qwen3-VL-2B. Each student uses a teacher trained from the same backbone.}
    \label{tab:gad_configuration}
    \small
    \setlength{\tabcolsep}{12pt}
    \renewcommand{\arraystretch}{1.15}
    \begin{tabular}{@{\hspace{8pt}}l >{\raggedright\arraybackslash}p{0.46\linewidth}@{\hspace{8pt}}}
        \toprule
        \textbf{Setting} & \textbf{Value} \\
        \midrule
        Student / teacher backbone & Qwen3.5-2B or Qwen3-VL-2B; matched within each pair \\
        Initialization & Student: base checkpoint; teacher: transcription SFT from the same base \\
        Teacher SFT data & 200,000 transcription-task examples \\
        Mixed dataset & 14,400 samples (perturbed + general) \\
        Mixture (general:perturbed) & 3:2 \\
        Learning rate / duration & $10^{-6}$ / 300 steps \\
        Input batch / PPO minibatch & 48 / 48; microbatch 1 per GPU \\
        Responses per input & 8 (384 per rollout batch) \\
        Rollout temperature / top-$p$ & 1 / 1; top-$k$ filtering disabled \\
        Teacher Top-$K$ / default coefficient & 32 / $\lambda=0.005$ \\
        Main-result coefficient & $\lambda$ \\
        Group controls & Disable distillation if $\max_i R_i\geq0.95$; attenuation $\kappa=3$ \\
        Teacher / student distillation temperature & 1 / 1 \\
        Student prompt / response limits & 8,192 / 8,192 tokens \\
        Maximum image pixels & 4,194,304 \\
        Hardware & 8 H100 GPUs \\
        \bottomrule
    \end{tabular}
\end{table}

\paragraph{Teacher input.}
The teacher receives reference text in a standalone heading-rewrite prompt, with no image. It changes existing heading prefixes, preserves all other characters, and outputs no outer Markdown fence. The same teacher prompt is used for offline signal analysis.

\paragraph{Training reward components.}
For text-perturbed documents, we use $\eta=0.5$ to give equal weight to edit similarity and perturbed-word recall. For output $y$ and GT $z$, the reward components in Section~\ref{sec:reward_function} are
\begin{equation}
\begin{aligned}
    R_{\mathrm{edit}}(y,z)
    &=1-\frac{D_{\mathrm{lev}}(y,z)}{\max(|y|,|z|)},\\
    R_{\mathrm{recall}}(y,z)
    &=\frac{\sum_w\min\!\left(c_{M(z)}(w),c_y(w)\right)}{|M(z)|}.
\end{aligned}
\label{eq:training_reward_components}
\end{equation}
Here, $D_{\mathrm{lev}}$ is the Levenshtein distance, and $|y|$ and $|z|$ are text lengths. $M(z)$ is the multiset of annotated perturbed words as printed; $c_{M(z)}(w)$ and $c_y(w)$ count exact occurrences of word $w$ in that multiset and the output. The sum ranges over distinct annotated perturbed words, and $|M(z)|$ counts all annotated occurrences. These training rewards are distinct from the benchmark metrics in Appendix~\ref{app:evaluation_metrics}.

\paragraph{Truncated FKL.}
For the vocabulary-truncated approximation, let $\mathcal K_{i,t}$ contain the teacher's Top-$K$ candidates, with $K=32$. We replace $d_{i,t}$ in Equation~\ref{eq:grpo-privileged-fkl} with
\begin{equation}
\begin{aligned}
    d^{\mathcal K}_{i,t}(\theta)
    &=\sum_{v\in\mathcal K_{i,t}}
    \pi_{\theta_{\mathrm T}}(v\mid x_{\mathrm T},y_{<t}^{(i)})\\
    &\quad\times\log\frac{
    \pi_{\theta_{\mathrm T}}(v\mid x_{\mathrm T},y_{<t}^{(i)})}
    {\pi_\theta(v\mid x,y_{<t}^{(i)})}.
\end{aligned}
    \label{eq:privileged-topk-fkl}
\end{equation}
The teacher probabilities retain their original full-vocabulary mass. Without renormalization, this truncated surrogate can be negative. The corresponding token loss is
\begin{equation}
    \ell^{\mathcal K}_{i,t}(\theta)
    =w_{i,t}(\theta)d^{\mathcal K}_{i,t}(\theta),
    \label{eq:gad_weighted_fkl}
\end{equation}
where $w_{i,t}(\theta)=\operatorname{sg}[\pi_\theta(b_{i,t}\mid x,y_{<t}^{(i)})]$ and $b_{i,t}$ is the teacher's Top-1 token. The weight uses the student's full-vocabulary probability of this teacher-preferred token and is detached; gradients pass only through the KL term. No additional rollout-ratio weight is applied to OPD. Teacher-choice comparisons use the same objective and group controls; component comparisons remove the indicated control in Table~\ref{tab:gad_controls_ablation}.

\section{Evaluation Metrics}
\label{app:evaluation_metrics}

We use benchmark normalization and matching rules for the metrics in Section~\ref{sec:eval_metrics}.

\paragraph{GlitchText.}
For $K$ annotated anomalies, $K_{\mathrm{vis}}$ counts predictions matching the visible perturbed form and $K_{\mathrm{orig}}$ counts restorations to the original form under official span alignment:
\begin{equation}
 \mathrm{Ident}=K_{\mathrm{vis}}/K,
 \qquad \mathrm{Cor}=K_{\mathrm{orig}}/K.
\label{eq:eval_glitch}
\end{equation}
Omissions and other errors count toward neither numerator. Table~\ref{tab:main_glitchtext} reports equal-weight means of the separate GlitchText-ZH and GlitchText-EN-Padding200 scores, expressed as percentages.

\paragraph{CHAOS-Bench.}
For perturbed words $\mathcal P_i$ and prediction $\hat y_i$ on page $i$, with $h$ indicating a case-insensitive whole-word match,
\begin{equation}
 \mathrm{MicroRecall}_{\mathrm{CHAOS}}
 =\frac{\sum_i\sum_{w\in\mathcal P_i}h(w,\hat y_i)}{\sum_i|\mathcal P_i|}.
\label{eq:eval_chaos}
\end{equation}
Each annotated target has equal weight across pages; unperturbed words are excluded.

\paragraph{General parsing.}
We report the official OmniDocBench v1.6 Overall score as a percentage.

\Needspace{8\baselineskip}
\section{Case Study}
\label{app:teacher_signal_cases}

\begin{figure}[!htbp]
    \centering
    \setlength{\abovecaptionskip}{4pt}
    \includegraphics[width=\linewidth]{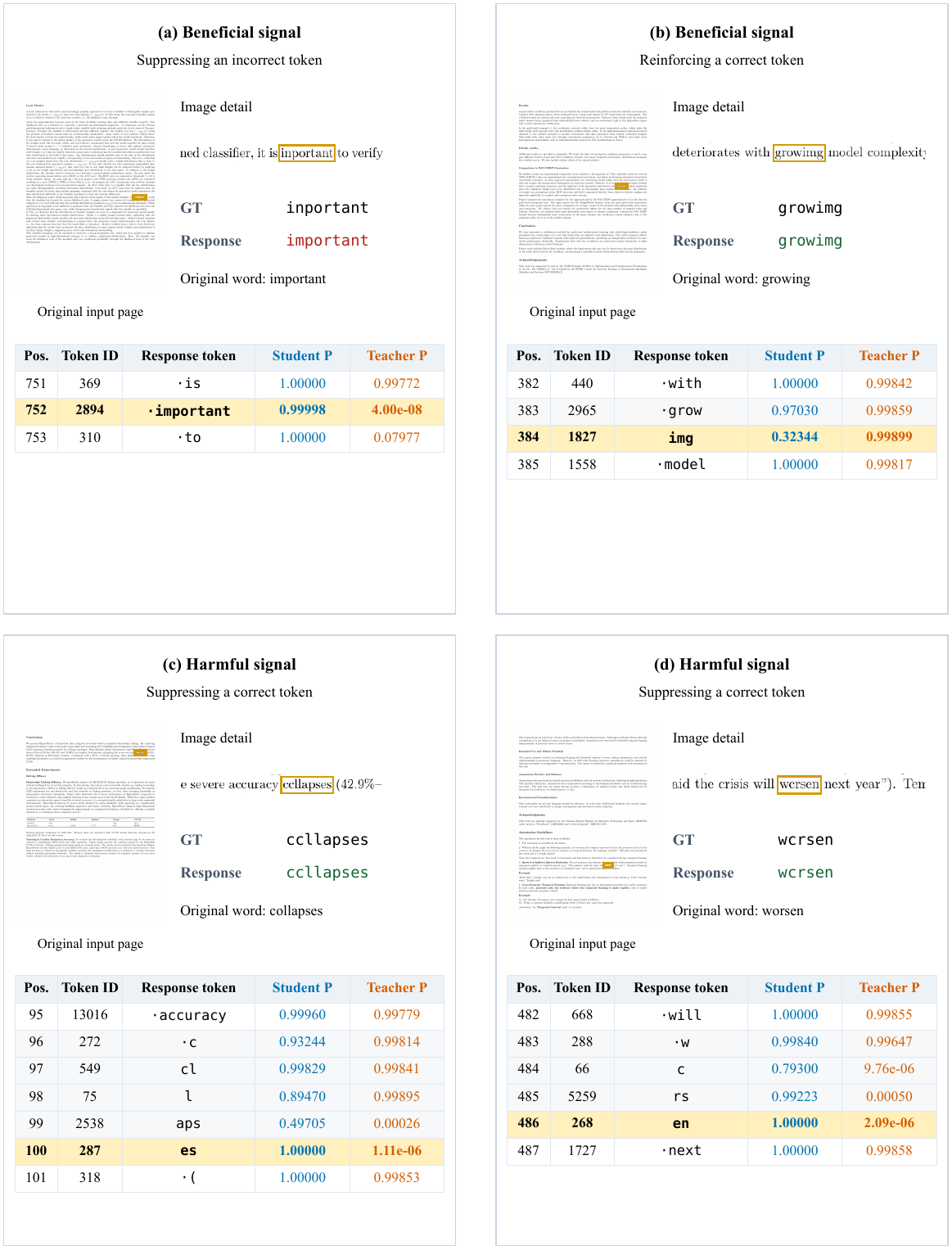}
    \caption{Beneficial and harmful teacher signals. Panels show the page, enlarged perturbed-word region, GT, student response, and token probabilities. Yellow rows mark focal tokens; \texttt{\textperiodcentered} denotes a space. Student and teacher score the same emitted token at the same prefix, conditioned on image and privileged text, respectively.}
    \label{fig:teacher_signal_cases}
\end{figure}

Figure~\ref{fig:teacher_signal_cases} illustrates both useful and conflicting guidance. In (a), the teacher suppresses an incorrect restoration to ordinary spelling; in (b), it reinforces a faithful perturbed-word token. In (c) and (d), it suppresses correctly transcribed perturbed-word-associated tokens. Correctness follows the printed GT, including intentional spelling perturbations. Thus privileged conditioning can produce both directionally beneficial and potentially conflicting signals, depending on token correctness and the direction of probability change.

\end{document}